%% file: root.tex
\documentclass[letterpaper, 10 pt, journal, twoside]{IEEEtran}  

\IEEEoverridecommandlockouts                                    

\usepackage{amsmath,amsfonts,amssymb}
\usepackage{algorithmic}
\usepackage[ruled,vlined,linesnumbered,boxed]{algorithm2e}

\let\origalg=\algorithm
\def\algorithm{\origalg\setlength{\commentWidth}{7cm}\DontPrintSemicolon\small\SetKwInput{KwParams}{Params}}
\makeatletter
\newcommand{\algrule}[1][.5pt]{\par\vskip.3\baselineskip\hrule height #1\par\vskip.3\baselineskip}
\makeatother
\newlength{\commentWidth}
\usepackage{array}
\usepackage[caption=false]{subfig}
\usepackage{textcomp}
\usepackage{stfloats}
\usepackage{url}
\usepackage{verbatim}
\usepackage{graphicx}
\usepackage{cite}
\usepackage{bm}
\usepackage{color}
\usepackage{xcolor}
\usepackage{multirow}
\usepackage{booktabs} 
\usepackage{tabularx}
\usepackage{lipsum}  
\usepackage[export]{adjustbox}
\usepackage{float}

\usepackage{gensymb}

\usepackage{tikz}
\usepackage{datetime}
\usetikzlibrary{shapes,arrows}

\makeatletter
\let\NAT@parse\undefined
\makeatother
\usepackage{hyperref}
\hypersetup{
    colorlinks=true,
    linkcolor=black,
    filecolor=black,      
    urlcolor=black,
    citecolor=black,
    pdftitle={Geometric Model Predictive Path Integral for Agile UAV Control with Online Collision Avoidance},
    }

\usepackage[normalem]{ulem}

\newcommand{\reffig}[1]{Fig.~\ref{#1}}

\newcommand{\refalg}[1]{Alg.~\ref{#1}}
\newcommand{\refsec}[1]{Sec.~\ref{#1}}
\newcommand{\reftab}[1]{Table~\ref{#1}}
\newcommand{\refeq}[1]{\eqref{#1}}

\SetKwInput{KwData}{Global}

\newcommand{\vect}[1]{\boldsymbol{\mathbf{#1}}}
\usepackage{physics}
\usepackage{siunitx}
\usepackage{dsfont}
\usepackage{nicefrac}

\title{Geometric Model Predictive Path Integral for Agile UAV Control with Online Collision Avoidance}

\author{Pavel Pochobradský, Ondřej Procházka, Robert Pěnička, Vojtěch Vonásek, Martin Saska%
\vspace{-2.2em}
\thanks{Manuscript received: October, 9, 2025; Revised January, 13, 2026; Accepted February, 8, 2026.}
\thanks{This paper was recommended for publication by Editor G. Loianno upon evaluation of the Associate Editor and Reviewers’ comments.}%
\thanks{
The authors are with the Multi-robot Systems Group, Faculty of Electrical
Engineering, Czech Technical University in Prague, Czech Republic (\protect\url{http://mrs.felk.cvut.cz/}).
This work has been supported by the Czech Science Foundation (GA\v{C}R) under research project No. 23-06162M, by the European Union under the project Robotics and advanced industrial production (reg. no. CZ.02.01.01/00/22\_008/0004590), and by CTU grant no SGS23/177/OHK3/3T/13.
}
\thanks{Digital Object Identifier (DOI): see top of this page.}
}

\usepackage[nolist]{acronym}

\begin{acronym}
  \acro{API}[API]{Application Programming Interface}
  \acro{CTU}[CTU]{Czech Technical University}
  \acro{DOF}[DOF]{degree-of-freedom}
  \acro{FOV}[FOV]{Field of View}
  \acro{GNSS}[GNSS]{Global Navigation Satellite System}
  \acro{GPU}[GPU]{Graphics Processing Unit}
  \acro{CNN}[CNN]{Convolutional Neural Network}
  \acro{GPS}[GPS]{Global Positioning System}
  \acro{GMPPI}[GMPPI]{Geometric Model Predictive Path Integral}
  \acro{MPPI}[MPPI]{Model Predictive Path Integral}
  \acro{IMU}[IMU]{Inertial Measurement Unit}
  \acro{LKF}[LKF]{Linear Kalman Filter}
  \acro{LTI}[LTI]{Linear time-invariant}
  \acro{LiDAR}[LiDAR]{Light Detection and Ranging}
  \acro{MAV}[MAV]{Micro Aerial Vehicle}
  \acro{NMPC}[NMPC]{Nonlinear Model Predictive Control}
  \acro{MPC}[MPC]{Model Predictive Control}
  \acro{PID}[PID]{Proportional Integral Derivative}
  \acro{LQR}[LQR]{Linear Quadratic Regulator}
  \acro{MRS}[MRS]{Multi-robot Systems Group}
  \acro{ROS}[ROS]{Robot Operating System}
  \acro{RTK}[RTK]{Real-time Kinematics}
  \acro{SLAM}[SLAM]{Simultaneous Localization And Mapping}
  \acro{UAV}[UAV]{Unmanned Aerial Vehicle}
  \acro{APF}[APF]{Artificial Potential Field}
  \acro{UGV}[UGV]{Unmanned Ground Vehicle}
  \acro{UKF}[UKF]{Unscented Kalman Filter}
  \acro{MOCAP}[MOCAP]{Motion Capture System}
  \acro{APF}[APF]{Artificial Potential Field}
  \acro{VFH}[VFH]{Vector Field Histogram}
  \acro{RL}[RL]{Reinforcement Learning}
  \acro{RMSE}[RMSE]{Root Mean Square Error}
  \acro{IL}[IL]{Imitation Learning}
  \acro{CCS}[CCS]{Constrained Covariance Steering}
  \acro{CVaR}[CVaR]{Conditional Value-at-Risk}
  \acro{PA-MPPI}[PA-MPPI]{Perception-Aware MPPI}
\end{acronym}

\begin{document}

\maketitle


\begin{abstract}
In this letter, we introduce \ac{GMPPI}, a sampling-based controller capable of tracking agile trajectories while avoiding obstacles.
In each iteration, \ac{GMPPI} generates a large number of candidate rollout trajectories and then averages them to create a nominal control to be followed by the controlled \ac{UAV}.
Classical \ac{MPPI} faces a trade-off between tracking precision and obstacle avoidance; high-noise random rollouts are inefficient for tracking but necessary for collision avoidance.
To this end, we propose leveraging geometric SE(3) control to generate a portion of \ac{GMPPI} rollouts.
To maximize their benefit, we introduce a \ac{UAV}-tailored cost function balancing tracking performance with obstacle avoidance.
All generated rollouts are projected onto depth images for collision avoidance, representing, to our knowledge, the first method utilizing depth data directly in a \ac{UAV} \ac{MPPI} loop.
Simulations show \ac{GMPPI} matches the tracking error of an obstacle-blind geometric controller while exceeding the avoidance capabilities of state-of-the-art planners and learning-based controllers.
Real-world experiments demonstrate flight at speeds up to 17 m/s and obstacle avoidance up to 10 m/s.
\end{abstract}


\vspace{-1.8em}
\section*{Supplementary Material}
{
  \footnotesize
  \vspace{-0.3em}
  \noindent \textbf{Video:} \url{https://youtu.be/HEo4MQNX6xc}\\
  \noindent \textbf{Code:} \url{https://github.com/ctu-mrs/gmppi}\\
  \vspace{-0.7em}
}

\acresetall



\vspace{-1.8em}

\section{Introduction\label{sec:introduction}}

\IEEEPARstart{A}{utonomous} \acp{UAV} are increasingly being deployed in missions requiring navigation in unknown cluttered environments. In various scenarios, such as search and rescue~\cite{schedl2021}, power line inspection~\cite{ollero2024}, and even digitization of historical monuments~\cite{petracek2024}, \acp{UAV} might need to perform complex maneuvers in unknown environments while avoiding obstacles.
Meanwhile, speed is an important criterion across use cases, allowing for better efficiency in the case of inspection and even helping save lives in the case of search and rescue.
Control and local trajectory planning for \acp{UAV} in such conditions is a challenging open problem.
Flying at higher speeds demands collision avoidance that supports rapid, agile maneuvers, but (small) UAVs are typically constrained by depth sensor range and computational capability.

The classical solution to real-time navigation in cluttered environments is modular: an environment map is built from sensor input, a global planner finds an obstacle-free trajectory, and a controller executes it.
In this approach, latency can accumulate, preventing flight at higher speeds~\cite{loquercio2021}.
Modularity often prevents the planner from considering full dynamics and the controller from accounting for obstacles, reducing agility.
Moreover, a controller that is not aware of obstacles might cause a crash even if the original planned trajectory was collision-free due to, e.g., a tracking error.
Recent learning-based methods aim to bypass the modular design by learning end-to-end mapping from sensor data to control commands~\cite{loquercio2021},~\cite{zhang2024}.
Although such methods avoid the segmentation challenges mentioned above, they require large training data and cannot be easily reused for different \acp{UAV}~\cite{scaramuzza2023}.
\begin{figure}[t!]
  \begin{center}
    \input{assets/tikz/intro_figure}
  \end{center}
  \vspace{-1.6em}
  \caption{
  Visualisation of the real-world experiment verifying the obstacle avoidance capability of the proposed controller.}
  \vspace{-1.8em}
  \label{fig:intro}
\end{figure}
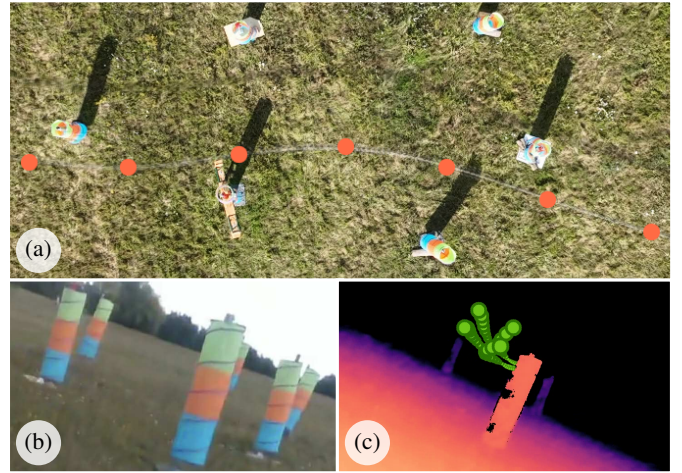

This letter proposes a novel \ac{UAV} controller which we call \ac{GMPPI}.
Based on the \ac{MPPI} method~\cite{minarik2024}, \ac{GMPPI} is a sampling-based controller utilizing a \ac{GPU} for parallel generation of a large number of candidate rollout trajectories in each iteration, resulting in obstacle avoidance and agile \ac{UAV} control capability.
Like learning-based methods~\cite{loquercio2021,zhang2024}, our approach unifies planning and control, but requires no training and is reusable across \acp{UAV} by modifying parameters.
Tuning existing \ac{MPPI} methods~\cite{minarik2024} requires balancing trajectory tracking precision, flight smoothness, and obstacle avoidance capability.
In particular, rollout trajectories generated by adding noise to a nominal trajectory do not represent an efficient way of following a reference trajectory.

Therefore, in addition to random rollouts, we propose using geometric SE(3) control~\cite{lee2010se3} to generate a portion of the GMPPI rollout trajectories, enabling precise reference tracking. Notably, even a small number of such SE(3)-based rollouts significantly improves trajectory tracking performance compared to prior MPPI approaches~\cite{minarik2024}. To maximize this benefit, we introduce a \ac{UAV}-tailored \ac{GMPPI} cost function, which enables us to retain the tracking capability of SE(3) control alongside the obstacle avoidance capability.
Furthermore, to the best of our knowledge, we introduce the first method utilizing depth sensor data directly in a \ac{UAV} \ac{MPPI}-based controller loop.
We also adopt a variable time-step sampling strategy, balancing the importance of near and distant trajectory segments while fully exploiting sensor range. Simultaneously, deterministic yaw control improves sampling performance without increasing the number of samples.

We show that in trajectory tracking tasks, the proposed \ac{GMPPI} controller achieves a tracking error similar to geometric SE(3) control~\cite{lee2010se3} and significantly smaller than an existing \ac{MPPI} controller~\cite{minarik2024}.
Through an ablation study, we highlight the contribution of each of the \ac{GMPPI} features to overall performance, namely integration of SE(3) rollouts, variation of noise and cost parameters with time step and dynamic rollout time step length.
In obstacle avoidance capability, \ac{GMPPI} exceeded the performance of the learning-based controllers~\cite{loquercio2021, zhang2024} and of the state-of-the-art obstacle-aware Bubble planner~\cite{ren2022}.
In real-world deployment, shown in \reffig{fig:intro}, the proposed controller is capable of avoiding obstacles at speeds of up to \SI{10}{\meter\per\second}.


\section{Related work}\label{sec:related_work}


\subsubsection{Trajectory Tracking\label{subsec:related_traj_track}}
Quadrotor trajectory tracking ranges from linear \ac{PID} to \ac{NMPC}~\cite{sun2022} and SE(3) control~\cite{lee2010se3}.
SE(3) control stands out due to its almost global exponential stability~\cite{lee2010se3}, enabling aggressive flight at the limit of dynamics as long as the reference trajectory is feasible. 
Its low computational overhead allows fast on-board processing and integration into the proposed controller, while its tracking capability makes it a suitable benchmark.


\subsubsection{Obstacle Avoidance\label{subsec:related_obs_avoid}}
The task of \acp{UAV} avoiding obstacles is generally handled as part of local trajectory optimization either by a local planner~\cite{ren2022} or a controller~\cite{garimella2017}
One of the oldest methods of real-time replanning and optimization of collision-free trajectories utilized \ac{APF}~\cite{khatib1985}, where an artificial force defined as the gradient of an \ac{APF} acts on the trajectory, the goal position has the lowest potential, and states with a possible collision have a high potential.
Alternatively, \ac{VFH} \cite{borenstein1991} and \ac{VFH}+ \cite{ulrich1998} discretize the possible trajectories into a polar histogram and balance path smoothness, distance to goal, and obstacle avoidance.
Both \ac{APF} and \ac{VFH}+ have been applied to \ac{UAV} obstacle avoidance in \cite{budiyanto2015} and \cite{fraundorfer2012}, respectively.

\ac{APF} and \ac{VFH}+ cannot take full advantage of the agility of some \acp{UAV}.
Directly finding a safe, e.g., time-optimal trajectory utilizing the full \ac{UAV} agility while avoiding obstacles is not computationally feasible in real time.
Instead of directly optimizing the trajectory, precomputed motion primitives can be used to build a feasible trajectory.
Such primitives can be state-based, control-based, or motion-based \cite{lopez2017}.
Improvements to motion primitive-based search algorithms for local obstacle-aware trajectory planning were developed in \cite{liu2018}.
To further decrease trajectory computation time and latency, the mapping step can be bypassed.

Methods such as \cite{lopez2017} or \cite{tordesillas2019} use most recent image data to generate local trajectories.
Despite integration of these methods into navigational pipelines and extensive testing \cite{Mohta2018ExperimentsIF}, they cannot utilize the full dynamic capabilities of agile \acp{UAV}.

As the quadrotor \ac{UAV} is differentially flat \cite{mellinger2011}, an alternative to motion primitive-based planning is to directly solve the trajectory optimization problem by creating polynomial trajectories \cite{wang2022}.
The method \cite{wang2022} has been applied along with a corridor planner in \cite{ren2022}, with performance exceeding some methods based on \ac{IL}~\cite{loquercio2021}.
Alternatively, obstacle avoidance can be included as a condition of trajectory optimization~\cite{garimella2017} in an adapted \ac{MPC} controller, though this is computationally demanding.

Although recent methods utilizing separate mapping, planning and control modules have significantly advanced agile flight capabilities \cite{ren2022}, alternatives mitigating latency have been explored, such as learning-based approaches that generate control commands from sensor inputs \cite{loquercio2021}, using either \ac{RL}~\cite{dai2020} or \ac{IL}~\cite{loquercio2021}.
Although these methods produce impressive results, they embed very little prior information about the systems they are controlling, treating them instead as black boxes.
The controller must therefore be retrained after any change to the parameters of the controlled UAV, such as mass.
Additionally, \ac{RL} and \ac{IL} methods have very low sample efficiency \cite{heeg2024}.
Recent research has suggested mitigating the low sample efficiency by using differentiable simulations~\cite{zhang2024}, but it does not mitigate the other drawbacks of learning-based methods.


\subsubsection{Model Predictive Path Integral\label{subsec:related_mppi}}

\ac{MPPI} is a variant of \ac{MPC} utilizing principles of Path Integral \cite{kazim2024} control.
While traditional \ac{MPC} solves local trajectory optimization using iterative methods, \ac{MPPI} uses a Monte Carlo sampling-based approach~\cite{williams201601}, allowing it to work with gradient-free and non-smooth cost functions.

Pure \ac{MPPI} control has a number of drawbacks.
First, a good initialization of the nominal commands is required.
For \ac{UAV} control, this can be solved by reusing the results of previous \ac{MPPI} iterations.
Second, disturbances that are not taken into account by the model used for path integration can cause issues with convergence.
This can be addressed by generating rollouts from the desired state and implementing an ancillary controller to follow this trajectory, effectively using the \ac{MPPI} only as a local planner and delegating the controller functionality \cite{Williams2018RobustSB}.

\ac{MPPI} has been explored for quadrotor control in both simulation~\cite{li2025} and real-world experiments~\cite{minarik2024}. Although a potential for trajectory tracking and obstacle avoidance has been demonstrated, obstacle avoidance was limited to hard-coded obstacles. Moreover, no configuration of \ac{MPPI} demonstrated the ability to combine effective obstacle avoidance with agile and smooth trajectory tracking at the level achieved by methods dedicated to each task individually.

More recently, \ac{PA-MPPI}~\cite{zhai2025pa-mppi} was introduced, which augments \ac{MPPI} with perception objectives for exploration when the goal is occluded. This extends \ac{MPPI} toward global navigation, while our approach focuses on local planning and control for agile flight with collision avoidance.


\section{Methodology\label{sec:methodology}}


This section begins with a brief overview of \ac{MPPI} control.
Then it details the architecture of the proposed \ac{GMPPI} controller, highlighting the integration of depth sensing, implementation of rollouts generated using an SE(3) controller, and the development of a custom cost function enabling agile and smooth flight as well as obstacle avoidance.

The base \ac{MPPI} terminology and notation shown in this section are adopted from~\cite{minarik2024}, where the complete method is presented.
\ac{MPPI}, when used as a controller, is capable of tracking trajectories while avoiding obstacles.
It is first initialized with a nominal control sequence $\vect{u}^{\mathrm{nom}} = \begin{bmatrix} \vect{u}^{\mathrm{nom}}_0, \ldots, \vect{u}^{\mathrm{nom}}_{N-1} \end{bmatrix}$, where the lower index $j$, $0 \le j < N$, indicates a value at the $j$th time step of a sequence (i.e., $u_j \in \vect{u}^{\mathrm{nom}}$). 
In each iteration, $K$ disturbance sequences $\delta{\vect{u}_j^k}$, each of length $N$, are sampled from a normal distribution with zero mean.
Rollout commands $\vect{u}_j^k$ and states $\vect{x}_{j+1}^k$ are then computed as
\begin{equation}
  \left.\begin{aligned}
    \delta{\vect{u}_j^k} & \in \mathcal{N}(0, \Sigma), \\
    \vect{u}_j^k & = \vect{u}^{\text{nom}}_j + \delta \vect{u}_j^k, \\
    \vect{x}_{j+1}^k & = \vect{x}_{j}^k + \vect{f}_{\text{RK4}}(\vect{x}_j^k, \vect{u}_j^k, \Delta t),
  \end{aligned} \hspace{0.5em}\right\}\hspace{0.5em}
  \begin{aligned}
    k & = 1, \dots, K,       \\
    j & = 0, \dots, N - 1.
  \end{aligned}
  \label{eq:rollout_computation}
\end{equation}

For \ac{UAV} control, the state is $\vect{x}=\begin{bmatrix}\vect{p}^T &\vect{v}^T & \vect{q}^T & \vect{\omega}^T\end{bmatrix}^T$, consisting of the position, velocity, and body rate vectors $\vect{p},\vect{v},\vect{\omega}\in\mathbb{R}^3$, respectively, and unit quaternion rotation on the rotation group $\vect{q} \in \mathbb{SO}(3)$.
Commands $\vect{u}=\begin{bmatrix}F_t & \vect{\omega}_c\end{bmatrix}$ consist of the total desired thrust $F_t\in\mathbb{R}^+$ and angular velocity $\vect{\omega}_c\in\mathbb{R}^3$. Commands are limited to $F_{t,\text{min}} \le F_t \le F_{t,\text{max}}$, %
$ \left|{\omega_{c,x}}\right| \le \omega_{\text{xy, max}}$, %
$ \left|\omega_{c,y}\right| \le \omega_{\text{xy, max}}$ %
and $ \left|\omega_{c,z}\right| \le \omega_{\text{z, max}}$ %
before being used to generate a rollout trajectory. The values of the time step $\Delta t$ used in \ac{GMPPI} are described in \refsec{subsec:dynamic_t}.
Dynamics of the \ac{UAV} are defined by
\begin{equation}
  \label{eq:drone_dynamics}
  \begin{aligned}
    \dot{\vect{p}} & = \vect{v},\text{ }\dot{\vect{q}} = \frac{1}{2} \vect{q} \odot \begin{bmatrix} 0 \\ \vect\omega \end{bmatrix},\text{ } \dot{\vect{\omega}} = \mathbf{J}^{-1}\left(\vect{\tau} - \vect{\omega}_c\times\mathbf{J}\vect{\omega}_c\right),\\
    \dot{\vect{v}} & = \frac{1}{m} \mathbf{R}(\vect{q})\left(\begin{bmatrix} 0 & 0 & F_t \end{bmatrix}^T - \vect{D}\mathbf{R}(\vect{q})^T\vect{v}\right) + \mathbf{g}, \\
  \end{aligned}
\end{equation}
with Runge-Kutta 4 being used in \refeq{eq:rollout_computation} for forward integration.
$\vect{R}(\vect{q})$ is a rotation matrix corresponding to the quaternion $\vect{q}$ and air resistance of the \ac{UAV} is approximated by a linear drag with coefficients $\vect{D} = \text{diag}\left(c_x, c_y, c_z\right)$.

Each rollout state sequence $\vect{x}^k$ is evaluated by a cost function to produce a cost $C^k$. Trajectories colliding with an obstacle and trajectories further from the reference receive higher costs.
The proposed \ac{GMPPI} cost function mapping rollout state sequences $\vect{x}^k$ to costs $C^k$ is presented in \refsec{subsec:cost_fn}.
Costs $C^k$ are used to compute rollout weights $w_k$ using
\begin{equation}
  \begin{aligned}
    \rho & = \min \{C^1, \dots, C^K\},
    &
    \tau_k & = -\frac{1}{\lambda} \left(C^k - \rho\right),
    \\
    \eta & = \sum_{k=1}^{K} \exp\left(\tau_k\right), 
    &
    w_k & = \frac{1}{\eta}\exp\left(\tau_k\right),
  \end{aligned}\label{eq:cost_2}
\end{equation}
with the parameter $\lambda$ controlling the degree to which a difference in cost between two trajectories causes a difference in their final weights.
Finally, the nominal control actions are set to the weighted average of the rollout commands
\begin{equation}
    \vect{u}^{\text{nom}}_j := \sum_{k=1}^{K} w_k \cdot \vect{u}_j^k\label{eq:u_update}
\end{equation}
and the first command $\vect{u}^{\text{nom}}_0$ is applied to the controlled \ac{UAV}.


\subsection{Depth Camera Integration\label{subsec:depth_cam}}
To the best of our knowledge, no \ac{MPPI} implementation has yet been integrated with a depth sensor for agile collision avoidance.
To minimize delay, we avoid map construction or heavy model processing~\cite{loquercio2021}. Instead, we use the available rollout states $\vect{x}^k_j = \begin{bmatrix} {\vect{p}^k_j}^T & {\vect{v}^k_j}^T & {\vect{q}^k_j}^T & {\vect{\omega}^k_j}^T \end{bmatrix}^T$.

\begin{figure}[bt]
  \begin{center}
    \includegraphics[width=1.0\linewidth]{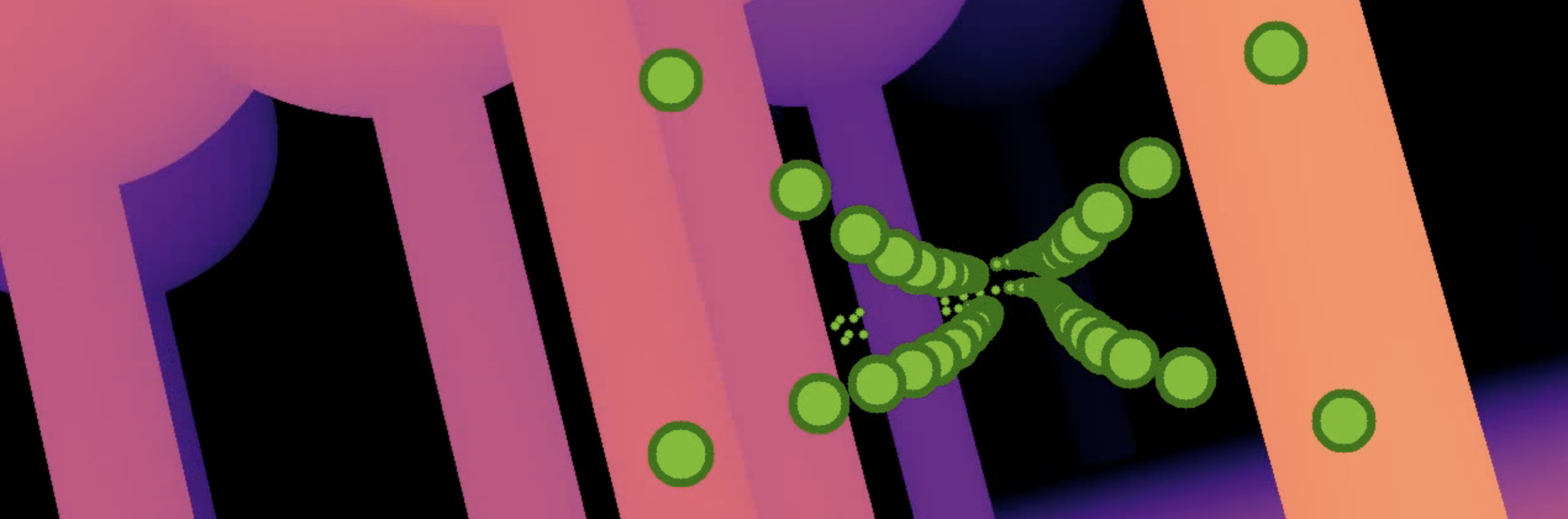}
  \end{center}
  \vspace{-1.5em}
  \caption{Depth-camera image with the projection of the front four corners of the \ac{UAV} collision box as defined in \refeq{eq:corners} in each of the positions in the nominal trajectory $\vect{p}^{\text{nom}}_j~\forall j \in (0, N+1)$, which is the result of applying the nominal control sequence $\vect{u}^{\text{nom}}_j~\forall j \in (0, N)$ to the UAV in its current state. } 
  \label{fig:depth_cam_img}
  \vspace{-0.4cm}
\end{figure}

The length $L$, width $W$ and height $H$ of the \ac{UAV} are used to define a set 
\begin{equation}
  \mathcal{H}^k_j = \left\{ \vect{p}^k_j + \frac{\epsilon}{2}\left. \begin{bmatrix} 
    \sigma_1L \\
    \sigma_2W \\
    \sigma_3H 
  \end{bmatrix} \right| \sigma_1,\sigma_2,\sigma_3 \in \{-1, 1\} \right\}
  \label{eq:corners}
\end{equation}
of all corner points of the \ac{UAV} shifted outwards by a safety multiplier $\epsilon>1$.
We project each of the corner points as well as the \ac{UAV} center point $\vect{p}_\mathrm{proj} \in \left(\mathcal{H}^k_j \cup \left\{ \vect{p}^k_j \right\}\right)$ onto the latest available depth image to determine if a collision occurs (see \reffig{fig:depth_cam_img}).
This approach eliminates nearly all sources of latency, with the exception of that introduced by the camera frame rate.

The sensor used in this work is a depth camera with an intrinsic matrix $\vect{K}$ that describes the transformation from the camera reference frame $C$ to the image reference frame $I$.
The camera is rigidly mounted to the \ac{UAV}, with the transformation from the body-fixed frame $B$ to $C$ described by a matrix $\vect{M}$.
The transformation from $B$ to the world frame $W$ at the time of the latest image being captured is described by $\vect{R}(\vect{q})_l$.

The position of each projected point $\vect{p}_\mathrm{proj}$ in the camera reference frame $C$ is obtained as
\begin{equation}
  {}^{C}\vect{p}_\mathrm{proj} = \left(\vect{R}(\vect{q})_l\vect{M}\right)^{-1} \vect{p}_\mathrm{proj}.
\end{equation}
This approach allows reusing a single image across multiple control iterations as it automatically compensates for the position and orientation shift of the \ac{UAV} since the last image was taken.
The distance of the rollout state position from the camera $\norm{{}^{C}\vect{p}_\mathrm{proj}}$ is compared to the distance $d_{px}$ measured by the camera at the pixel coordinates ${}^{I} \vect{p}_\mathrm{proj} = \vect{K}{}^{C} \vect{p}_\mathrm{proj}$, where the rollout position is projected on the depth image.

The existence of a collision is determined using
\begin{equation}
  Col(\vect{p}_\mathrm{proj}) = \mathlarger{\mathlarger{\mathlarger{\mathds{1}}}}_{\norm{{}^{C} \vect{p}_\mathrm{proj}} \in [d_{px}, d_{px} + d_a]}
  \label{eq:collision_check}
\end{equation}
is defined, which returns $1$ if the distance of the relevant trajectory point $\vect{p}_{\text{proj}}$ from the camera is similar to the distance $d_{px}$ measured by the camera at the pixel where $\vect{p}_{\text{proj}}$ would appear on the depth image.
Instead of assuming all space behind an obstacle is occupied, a depth $d_a$ is assumed to allow the controller to optimistically plan a return path to the reference trajectory even if the path is not fully visible yet.
If a projection falls outside of the visible area, the nearest pixel of the depth image is used to estimate the presence of obstacles that are partially outside of the field of view of the camera.

\subsection{Dynamic Rollout Time Steps\label{subsec:dynamic_t}}

Small \acp{UAV} are limited in terms of the on-board hardware, which in turn limits the computational power.
This restricts the number of rollouts and the number of steps in each rollout.

To ensure that the sensor range, labelled $s$, is fully utilized at any flight speed, a vector $\vect{n} = \begin{bmatrix} n_0, \ldots, n_{N-1} \end{bmatrix}$ is defined, where each element $n_j$ acts as a time step multiplier at the $j$th step of each rollout. 
The corresponding time step is calculated as $\Delta{t}_j = n_j\Delta{t}_0$, where $\Delta{t}_0=0.01\si{s}$ is the base controller update period.

\begin{figure}[bt]

  \begin{center}
    \includegraphics[width=0.9\linewidth]{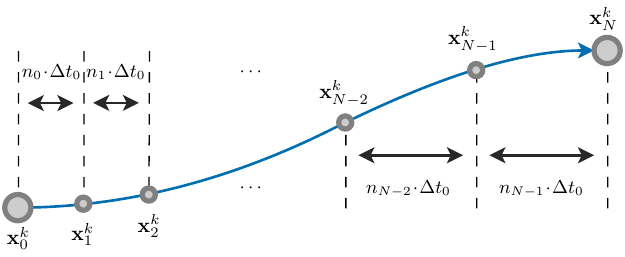}
  \end{center}
  \vspace{-1.5em}
  \caption{Illustration of the layout of the rollout states $\vect{x}^k_j$ in time. Earlier steps are shorter to ensure precision, while later steps are longer and their length is dynamically adjusted to use the full range of the available depth sensor.}
  \vspace{-0.4cm}
  
  \label{fig:timesteps}
\end{figure}

The first $M$ multipliers are fixed at smaller values represented by $n_{\text{near}}$ to maintain good simulation precision near the current state.
The remaining $N - M$ multipliers are set based on the average speed of the \ac{UAV} across the nominal state sequence $v^{\text{nom}}_{\text{avg}}$ and the sensor range $s$ such that
\begin{equation}
n_{\mathrm{far}} = \frac{\frac{s}{v_{\text{avg}}^{\text{nom}}\Delta{t}_0} - \sum_{j=0}^{M}{n_j}}{N-M},
  \label{eq:n_eq}
\end{equation}
which ensures that the range $s$ of the used sensor is fully utilized. The vector \vect{n} has values
\begin{equation}
  \begin{aligned}
    n_{0...M-1} = n_{\text{near}},
    \hspace{16pt}
    n_{M...N-1} = n_{\text{far}}
  \end{aligned}
\end{equation}
and the layout of a typical rollout is illustrated by \reffig{fig:timesteps}.

Aside from the sensor usage benefits, shorter time steps at the beginning of the rollout increase the importance of near-term flight precision and thus stability of \ac{GMPPI} in hover.

\subsection{Deterministic Yaw Control\label{subsec:yaw_ctrl}}
A key factor influencing performance of \ac{MPPI}-based control
is the density of the simulated rollout trajectories in the space of all possible rollout trajectories.
To increase this density without adding more rollouts, a dimension can be removed from the space.

This can be achieved by modifying the existing sampling-based \ac{MPPI}~\cite{minarik2024} rollout generation method to control the yaw axis rotation deterministically.
A proportional controller has been chosen for simplicity.
It follows the reference heading $\vect{h}^{\text{ref}}$ by computing
\begin{equation}
\vect{\omega}_z = k_{\text{MPPI},z}\angle(\vect{h}, \vect{h}^{\text{ref}}) + \vect{\omega}_{z,\text{ref}},
\end{equation}
where $k_{\text{MPPI},z}$ is the proportional feedback gain and $\angle(\cdot,\cdot)$ denotes the angle between two vectors.


\subsection{Geometric Control in Candidate MPPI Trajectories\label{subsec:se3}}

In \ac{MPPI}, the rollout command sequences $\vect{u}^k_j$ are generated by adding random noise $\delta{\vect{u}_j^k}$ to the nominal command sequence~$\vect{u}^{\text{nom}}_j$.
Such rollouts explore the space for a collision-free trajectory, but are not efficient at precisely following reference trajectories.
This motivates the search for rollout methods, which would allow precise trajectory tracking.
One such option is a geometric SE(3) controller.

The geometric SE(3) controller~\cite{lee2010se3} is a theoretically proven and practically validated method capable of following reference trajectories precisely and in a computationally efficient manner.
One of the practical downsides of this method is the tuning process.
It requires precise selection of gains, which often need to vary not just based on the controlled \ac{UAV}, but also the specific trajectory to be followed.

\ac{MPPI} and SE(3) control methods have complementary strengths and weaknesses.
Using the SE(3) controller as a rollout generation method produces rollouts capable of tracking a reference trajectory while the generation remains computationally feasible.
The remaining sampling-based rollouts can then focus on obstacle avoidance and exploration, as they no longer need to be used for following the reference trajectory.

\begin{figure}[t]
  \begin{center}
    \includegraphics[width=1.0\linewidth]{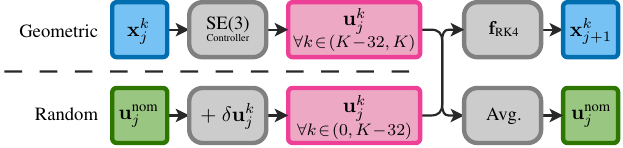}
  \end{center}
  \vspace{-1.5em}
  \caption{Combination of geometric and random rollouts. A total of $K$ rollout sequences are simulated, of which $K - 32$ are random and $32$ use geometric control. All rollout commands are used to compute the next nominal command sequence $\vect{u}^{\text{nom}}_j$ as well as the next set of rollout states $\vect{x}^{k}_{j+1}$.}
  \label{fig:se3_rollout}
  \vspace{-0.5cm}
\end{figure}

Moreover, because \ac{MPPI} already runs hundreds of parallel rollouts, more than a single SE(3) rollout may be added.
The amount of rollouts generated using SE(3) is labelled $K_{\text{SE3}}$, with the other $K - K_{\text{SE3}}$ remaining as before.
In practice, rollouts run on a \ac{GPU} in warps of $32$, so $K_{\text{SE3}} = 32$ is chosen, with the split between random and geometric rollouts illustrated in \reffig{fig:se3_rollout}.
Each of the SE(3) rollouts uses different parameters and multiple SE(3) configurations are therefore tested simultaneously, with the best configuration being selected by the proposed \ac{GMPPI} cost function.
Little benefit would arise from increasing or tuning the $K_{\text{SE3}}$ parameter, as the majority of the rollouts must remain sampling-based to preserve the ability to avoid obstacles.
The resulting controller can precisely track reference trajectories and avoid small unmapped obstacles.

To use a different geometric controller configuration in each rollout, the noise ${\delta{\vect{k}^k_{\text{SE3}}} \in \mathcal{N}(0, \Sigma_{\text{SE3}})}$ is sampled from a normal distribution with zero mean and a covariance matrix $\Sigma_{\text{SE3}}$. 
This noise is then added to a vector of \ac{GMPPI} parameters specifying a base geometric controller configuration ${\vect{k}_{\text{SE3}} = \begin{bmatrix} k_{pxy} & k_{pz} & k_{vxy} & k_{vz} & k_{rxy} & k_{rz} \end{bmatrix}}$ creating a new configuration specific to the rollout at index $k$ and defined by the parameters
\begin{equation}
  \vect{k}^k_{\text{SE3}} = \vect{k}_{\text{SE3}} + \delta{\vect{k}^k_{\text{SE3}}}.
\end{equation}
The values of the vector $\vect{k}^k_{\text{SE3}}$ are then used as parameters of the SE(3) controller~\cite{lee2010se3}.

\subsection{Cost Function\label{subsec:cost_fn}}

\ac{MPPI} computes a nominal trajectory as a weighted average of rollout trajectories (where higher weights $w_k$ are assigned to rollout trajectories with a better, i.e. lower cost, behaviour \refeq{eq:cost_2}).
Lower cost $C^k$ is assigned to trajectories conforming the objective of the control, while higher costs are assigned to trajectories that exhibit unwanted characteristics such as excessive maneuvers or a large tracking error.
The proposed controller has three goals, which are precise trajectory tracking, smooth flight, and obstacle avoidance, which always takes precedence.
In \ac{GMPPI}, the cost function selects whether the geometric or sampling-based rollouts should be prioritized.

Each unwanted characteristic is penalized by one or more cost components, where a measure of the unwanted characteristic is multiplied by a cost coefficient.
These coefficients are parameters of \ac{GMPPI} and they are dependent on the time index $j$ in the rollout, which allows prioritizing precise trajectory tracking and smooth flight early in the rollout and exploration later in the rollout.

\subsubsection{Tracking Precision \label{subsec:track_precision}}

To enable precise trajectory tracking, the errors in position, velocity, orientation, and angular velocity relative to their reference values are computed as
\begin{equation}
  \begin{aligned}
    e_p & = \norm{\vect{p}^k_j - \vect{p}^{\text{ref}}_j}_2, &
    e_v & = \norm{\vect{v}^k_j - \vect{v}^{\text{ref}}_j}_2, \\ 
    e_q & = d_{\vect{q}}(\vect{q}^k_j, \vect{q}^{\text{ref}}_j), &
    e_{\omega} & = \norm{\vect{\omega}^k_j - \vect{\omega}^{\text{ref}}_j}_2, 
  \end{aligned}
  \label{eq:pvqo_cost}
\end{equation}
where $d_{\vect{q}}(\vect{q}_1, \vect{q}_2)$ is an approximation of orientation difference adopted from~\cite{minarik2024} and defined as
\begin{equation}
  d_{\vect{q}}(\vect{q}_1, \vect{q}_2) = 1 - \langle\vect{q}_1, \vect{q}_2\rangle^2.
\end{equation}
The errors are multiplied by the position $c_j^p$, velocity $c_j^v$, orientation $c_j^q$ and angular velocity $c_j^\omega$ coefficients respectively.
The coefficients can vary with the time index $j$ in the rollout.
This allows assigning a lower cost to those rollout trajectories that deviate from the reference trajectory only in later steps of the rollout simulation, which promotes exploration and reallocates precision emphasis toward earlier timesteps of the simulated rollout trajectories, allowing the controller to maximize the use of the precision of the SE(3) rollouts.
This increases trajectory tracking precision, as well as smoothness.
Using the Euclidean norm for position instead of its square requires computing a square root but prevents the creation of an erratic nominal trajectory in cases where a larger deviation from the reference trajectory is necessary to avoid an obstacle.

\subsubsection{Smoothness \label{subsec:smoothness}}

To promote smooth flight, two cost components are added.
First, cost is added for jerk $\norm{\vect{\dot{a}}}$, which is larger than a multiple $t_j$ of the jerk present in the reference trajectory $||\vect{\dot{a}}^{\text{ref}}||$.
The excess jerk quantity
\begin{equation}
  e_j = \text{max}\left(||\vect{\dot{a}}^k_j|| - t_j||\vect{\dot{a}}^{\text{ref}}||, 0 \right)
  \label{eq:jerk_cost}
\end{equation}
is introduced and multiplied by a coefficient $c_j^j$.
This punishes any excess control input based on how much it disturbs the smoothness of the flight, while still allowing sharp trajectories to be followed precisely.
This is a key feature allowing the controller to use the full potential of agile SE(3)-generated rollout trajectories if the reference trajectory contains sharp maneuvers. The reference and rollout jerk values are calculated using a finite-difference approximation from acceleration data. 
Noise amplification concerns do not apply, as both the reference and rollout trajectories are noise-free.

Second, a cost is added for any difference in position 
\begin{equation}
    e_s=\norm{\vect{p}_j^k-\vect{p}^\mathrm{nom}}
    \label{eq:smooth_cost}
\end{equation}
between the rollout and the nominal trajectory.
This is multiplied by a coefficient $c_j^s$ and prevents the trajectory and therefore the control input from changing excessively between successive controller runs.

\subsubsection{Obstacle Avoidance \label{subsec:obstacle_avoidance}}

Obstacle avoidance capability is introduced by assigning cost to any rollout that collides with an obstacle. 
The set of points $\mathcal{H}^k_j \cup \left\{ \vect{p}^k_j \right\}$, defined in \refsec{subsec:depth_cam}, is a collision box for the \ac{UAV} and the function $Col(\vect{p}_\mathrm{proj})$, defined in \refeq{eq:collision_check}, returns $1$ when a collision is detected.
When a collision occurs at step $j$ of a rollout trajectory, the cost is scaled by $(N - j)$ so that collisions occurring earlier in the rollout are penalized more heavily than those further in the future. The obstacle cost is therefore defined as
\begin{equation}
  e_{\text{obs}} = (N - j)\sum\nolimits_{\vect{p}_\mathrm{proj} \in \left(\mathcal{H}^k_j \cup \left\{ \vect{p}^k_j \right\}\right)}Col(\vect{p}_\mathrm{proj}),
  \label{eq:obs_cost}
\end{equation}
and is further weighted by a coefficient $c^{\text{obs}}_j$.
This formulation provides a graded penalty rather than a binary collision/no-collision outcome, which allows for a more informed selection of candidate trajectories.

\subsubsection{Resulting Cost Function \label{subsec:resulting_cost_func}}

The combined cost for a single point in a single rollout is defined as
\begin{equation}
  \begin{aligned}
    \vect{c}_j & = \begin{bmatrix}
    c_j^p & c_j^v & c_j^q & c_j^\omega & c_j^j & c_j^s & c_j^{\text{obs}}
    \end{bmatrix}, \\
    \vect{e}_j^k & = \begin{bmatrix}
      e_p & e_v & e_q & e_\omega & e_j & e_s & e_{\text{obs}}
    \end{bmatrix}^T, \\
    C^k_j & = \vect{c}_j\cdot\vect{e}_j^k,
  \end{aligned}
  \label{eq:cost_function}
\end{equation}
resulting in the overall cost of a single rollout $k$ to be
\begin{equation}
  C^k = \sum\nolimits_{j=1}^{N} C^k_j.
\end{equation}


The entire \ac{GMPPI} algorithm is summarized in \refalg{alg:GMPPI}.
On lines 1--3, nominal and reference commands are resampled using the dynamic timesteps.
The algorithm then creates $K$ rollouts in parallel.
On lines 4--11, $K_{\text{SE3}}$ geometric rollouts are created by selecting the SE(3) parameters to use and then running the simulation.
On lines 12--16, random rollouts with proportional yaw control are calculated.
All rollouts are assigned a cost on lines 17--18, and finally, the new nominal command is created as a weighted average of rollouts on line~19.
The first nominal command is used by the \ac{UAV}.

{
\footnotesize
\begin{algorithm}
    \footnotesize
  \SetInd{0.5em}{0.5em}
  \SetKwInput{KwParams}{Params}
  \KwIn{Current state estimation $\hat{\vect{x}}$, Nominal command sequence $\vect{u}^{\text{nom}}_{\text{in}}$, Reference command sequence $\vect{u}^{\text{ref}}_{\text{in}}$, Depth Image}
  \KwParams{Number of rollouts $K$, Number of geometric rollouts $K_{\text{SE3}}$, Rollout length $N$, Rollout noise $\Sigma$, SE3 Parameter noise $\Sigma_{\text{SE3}}$}
  \algrule
  $\vect{\Delta t}$ = \textit{ComputeTimesteps()} \;
  $\vect{u}^{\text{nom}}$ = \textit{ResampleNominalCommands}($\vect{u}^{\text{nom}}_{\text{in}}$, $\vect{n}$) \;
  $\vect{u}^{\text{ref}}$ = \textit{ResampleReferenceCommands}($\vect{u}^{\text{ref}}_{\text{in}}$, $\vect{n}$) \;
  \vspace{4pt}
  \tcp{Simulate $K$ rollouts (Parallel on GPU)\hfill}
  \For(){$k$ = 1, \dots, $K$}{
    $\vect{x}_0^k = \vect{\hat{x}}$ \;
    \If(\tcp*[f]{\makebox[3cm]{SE(3) Rollout\hfill}}){$k < K_{\text{SE3}}$ } {
      $\delta\vect{k}^k_{\text{SE3}} \sim \mathcal{N}(0, \Sigma_{\text{SE3}})$ \;
      $\vect{k}^k_{\text{SE3}} = \vect{k}_{\text{SE3}} + \delta{\vect{k}^k_{\text{SE3}}}$ \;
      \For{$j$ = 0, \dots, $N$}{
        $\vect{u}^k_j$ = \textit{SE3Command($\vect{x}^k_j$, $\vect{k}^k_{\text{SE3}}$)} \;
        $\vect{x}_{j+1}^k$ = $\vect{x}_j^k + \vect{f}_{\text{RK4}}(\vect{x}_j^k, \vect{u}_j^k, \Delta{t}_j) $ \;
      }
    }
    \Else(\tcp*[f]{\makebox[3cm]{Random Rollout\hfill}}){
      \For{$j$ = 0, \dots, $N$}{
        $\delta\vect{u}^k_j \sim \mathcal{N}(0, \Sigma)$ \;
        $\vect{u}^k_j = \vect{u}^{\text{nom}} + \delta\vect{u}^k_j$ \;
        $\vect{\omega}_z = k_{\text{MPPI},z}\angle(\vect{h}, \vect{h}^{\text{ref}}) + \vect{\omega}_{z^{\text{ref}}}$
        $\vect{x}_{j+1}^k$ = $\vect{x}_j^k + \vect{f}_{\text{RK4}}(\vect{x}_j^k, \vect{u}_j^k, \vect{\omega}_z, \Delta{t}_j) $ \;
      }
    }
    $C^k_j$ = \textit{CalculateCost}($c^k_j$, $\vect{x}_{j+1}^k$, $\vect{x}_{j+1}^{\text{ref}}$, Depth Image) \;
    $C^k = \sum_{j=0}^{N}C^k_j$ \;
  }
  $\vect{u^{\text{nom}}}$ = \textit{AverageWeighedCommands}()\tcp*[f]{Eq. (\ref{eq:u_update}}) \;

  \Return $\vect{u}^{\text{nom}}_0$
  \caption{Single iteration of Geometric MPPI}
  \label{alg:GMPPI}
\end{algorithm}
}


\vspace{-1em}
\section{Results\label{sec:results}}


This section presents the results of conducted experiments.
First, simulations verifying the ability to track agile trajectories were conducted, including an ablation study showing the benefit of each feature described in \refsec{sec:methodology}.
Next, obstacle avoidance capability has been shown in simulation and compared against three existing methods.
Finally, the results of real-world experiments are presented.

All simulated experiments were conducted on a laptop with an Intel Core i7-1165G7 CPU and an NVIDIA GeForce MX450 \ac{GPU}, which the proposed controller utilizes for parallel rollout simulation.
Real-world experiments utilized a Jetson Orin NX high-level computing unit with an 8-core Arm Cortex-A78AE CPU and an Ampere \ac{GPU} with \ac{GMPPI} controller outputs fed into a PX4 low-level flight controller.
An Intel RealSense D435i depth camera set to a resolution of 640x480 px was used.
All reference trajectories were generated either by an MPC-based tracker~\cite{baca_mrs_2021} or by a polynomial trajectory planner~\cite{wang2021}.


\subsection{UAV \& Controller Parameters \label{sec:result_params}}

The \ac{UAV} mass $m$, the arm length $l$, the inertia matrix $\mathbf{J}$ and the rotor torque constant $c_{tf}$ of the simulated \ac{UAV} are shown in \reftab{tab:uav_params}. The parameters of the \ac{UAV} used for real-world experiments were similar. The \ac{GMPPI} parameters along with control input limits for both the minimum and maximum thrust $F_{t, \text{min}}, F_{t, \text{max}}$ as well as the maximum absolute angular rate $\omega_{\text{max}}$ are shown in \reftab{tab:params_limits}.
\begin{table}[b]
  \vspace{-1.1em}
  \scriptsize
  \centering
  \caption{UAV Parameters used for all simulated runs}
  \vspace{-1.2em}
  \label{tab:uav_params}
  \begin{tabular}{l c | l c | l c}
    \toprule
    \multicolumn{2}{c|}{Model Parameters} & \multicolumn{2}{c|}{Dimensions} & \multicolumn{2}{c}{Drag Coefficients} \\
    \midrule
    $m$ \hfill $[\si{\kg}]$ & $1.21$ & $L$ \hfill $[\si{\m}]$ & $0.35$ & $c_x$ & $0.28$ \\
    $l$ \hfill $[\si{\m}]$ & $0.15$ & $W$ \hfill $[\si{\m}]$ & $0.35$ & $c_y$ & $0.35$ \\
    $c_{tf}$ \hfill $[\si{\m}]$ & $0.012$ & $H$ \hfill $[\si{\m}]$ & $0.215$ & $c_z$ & $0.7$ \\
    \midrule
    \multicolumn{2}{l}{Inertia Matrix} & \multicolumn{1}{c}{$\mathbf{J}$ \hfill $[\si{\g\m^2}]$} & \multicolumn{3}{l}{$\text{diag}(\left[\begin{smallmatrix} 7.06 & 7.06 & 13.6 \end{smallmatrix}\right])$} \\
    \bottomrule
  \end{tabular}

\end{table}

For obstacle avoidance in simulation, a stereo depth camera with a range of ${s = 13~\si{\m}}$ is mounted to the \ac{UAV}, although the rollout length is limited to $\SI{10}{\m}$.
The controller treats every visible obstacle as occupying ${d_a = \SI{2.0}{\m}}$ along the viewing ray before free space is presumed.
All experiments had the camera tilted up by a fixed angle to allow full use of the sensor range. Parameters are shown in \reftab{tab:cam_tilts}.

\begin{table}
  \scriptsize
  \centering
  \caption{Controller Parameters and Limits\label{tab:params_limits}}
  \vspace{-1.2em}
  \begin{tabular}{l c | l c | l c}
    \toprule
    \multicolumn{2}{c}{MPPI Parameters} & \multicolumn{2}{c}{SE3 Parameters} & \multicolumn{2}{c}{Control Limits} \\
    \midrule
    $K$ & $768$ & $k_p$ & $\left[\begin{smallmatrix} 6.0 & 6.0 & 15.0 \end{smallmatrix}\right]$ & $F_{t,\text{min}}$ \hfill $[\si{\N}]$ & $0.46$ \\
    $K_{\text{SE3}}$ & $32$ & $k_v$ & $\left[\begin{smallmatrix} 4.0 & 4.0 & 8.0 \end{smallmatrix}\right]$ & $F_{t,\text{max}}$ \hfill $[\si{\N}]$ & $20.6$ \\
    $N$ & $30$ & $k_r$ & $5.0$ & $\omega_{\text{xy, max}}$ \hfill $[\si{\text{rad}\s^{-1}}]$ & $10.0$ \\
    $k_{\text{MPPI},z}$ & $2.0$ & & & $\omega_{\text{z, max}}$ \hfill $[\si{\text{rad}\s^{-1}}]$ & $2.0$ \\
    \bottomrule
  \end{tabular}
  \vspace{-1.1em}
\end{table}

\begin{table}
  \scriptsize
  \centering
  \caption{Relationship of Speed and Camera Tilt}
  \label{tab:cam_tilts}
  \vspace{-1.2em}
  \begin{tabular}{l | c | c | c | c | c | c | c | c}
      \toprule
      Speed $[\si{\m\per\s}]$ & 3 & 5 & 7 & 9 & 10 & 11 & 12 & 13 \\
      \midrule
      Tilt $[\text{deg}]$ & 8 & 10 & 16 & 22 & 22 & 27 & 27 & 30 \\
      \bottomrule
  \end{tabular}
  \vspace{-1.3em}
\end{table}

\begin{figure}[t]
  \centering
  \subfloat [Cost parameters $c_j^p, c_j^v, c_j^q, c_j^\omega$ and $c_j^j$ used in \refeq{eq:cost_function}.\label{fig:costs_graph}] {
    \includegraphics[width=0.44\linewidth]{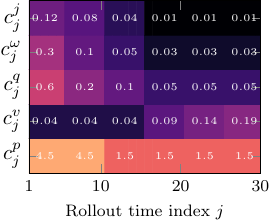}
  }
  \hspace{0.1cm}
  \subfloat [Noise parameters for the angular velocities $\sigma_{\omega_x}$, $\sigma_{\omega_y}$ and thrust $\sigma_{F_t}$.\label{fig:noise_graph}] {
    \includegraphics[width=0.46\linewidth]{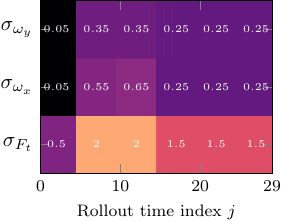}
  }
  \caption{\ac{GMPPI} cost and noise parameters used in all experiments. A decrease in $c_j^p$, $c_j^q$ and $c_j^\omega$ promotes exploration. An increase in $c_j^v$ prevents unnecessary flight direction changes. Higher noise in the middle of rollouts promotes exploration.\vspace{-2.5em}}
  \label{fig:noise_and_cost_graph}
\end{figure}

The relative size of cost parameters from \refeq{eq:cost_function}, shown in \reffig{fig:costs_graph}, $c_j^p$, $c_j^v$, $c_j^q$ and $c_j^\omega$ is based on an existing \ac{MPPI} implementation~\cite{minarik2024}.
They are, however, also dependent on the rollout index $j$.
$c_j^p$, $c_j^q$ and $c_j^\omega$ decrease with $j$ to enable exploration, while $c_j^v$ increases to prevent unnecessary flight direction changes while exploring.
$t_j = 1.4$ is constant.
Finally, the obstacle cost is $c_j^{\text{obs}} = 1000$.
This is applied in~\refeq{eq:obs_cost} so that avoiding collisions becomes the highest priority.

As with the cost parameters, the noise covariance matrix $\Sigma = \text{diag}\left(\begin{bmatrix} \sigma_{F_t} & \sigma_{\omega_x} & \sigma_{\omega_y} \end{bmatrix}\right)$, which is defined in \reffig{fig:noise_graph}, is dependent on the rollout time index $j$.
The relative size of $\sigma_{F_t}, \sigma_{\omega_x}$ and $\sigma_{\omega_y}$ is based on a previous \ac{MPPI} implementation~\cite{minarik2024}.
The noise is reduced in the first part of the rollout to prevent noise from being applied to the \ac{UAV} and also later on in the rollout where timesteps are longer to prevent erratic behaviour of the rollout trajectories.
Rotation around the yaw axis is controlled separately, as described in \refsec{subsec:yaw_ctrl}.


\subsection{Reference Tracking}\label{subsec:ref_tracking}

\begin{figure}
  \centering
  \subfloat [\textit{Figure~8}\label{fig:ref_fig8}] {
    \includegraphics[width=0.44\linewidth]{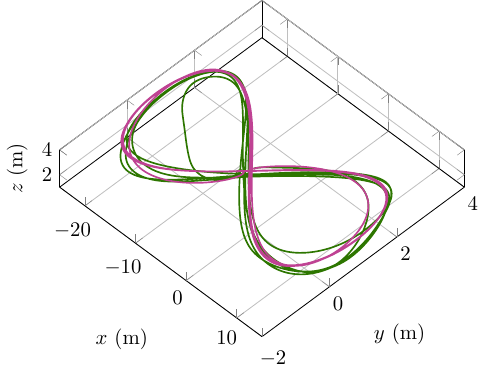}
  }
  \subfloat [\textit{Hypotrochoid}\label{fig:ref_hypo}] {
    \includegraphics[width=0.44\linewidth]{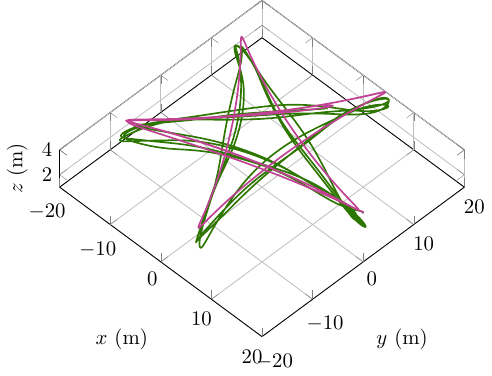}
  }
  \caption{Trajectories followed by \ac{GMPPI} shown in green and the corresponding reference trajectories shown in red.}
  \label{fig:ref_trajs}
  \vspace{-1.2em}
\end{figure}

To test the ability of the \ac{GMPPI} controller to follow reference trajectories, three trajectories were generated: a hover trajectory and two agile trajectories shown in \reffig{fig:ref_trajs}.
All trajectories specify the heading of the \ac{UAV} to be equal to the projection of the velocity vectors onto the horizontal plane to allow collision avoidance using a front-mounted stereo camera.

The \ac{GMPPI} controller set up with parameters from \refsec{sec:result_params} is compared to an existing standalone implementation of MPPI~\cite{minarik2024} and a standalone implementation of the SE(3) controller~\cite{lee2010se3}.
Three additional configurations of \ac{GMPPI} are included to verify the contribution of each of the key features to the overall performance of the controller.
The ``no SE(3)'' variant has a configuration modified with $K_{SE3} = 0$, disabling the geometric rollouts.
The ``const $\Delta{t}$'' variant includes $\vect{n} = [(n_{\text{exp}})_{\times N}]$, 
which equalizes the timestep lengths in all time steps of all rollouts.
Finally, the ``const noise'' variant has all costs and noise parameters set to be constant for all time steps of all rollouts.

The results are shown in \reftab{tab:traj_results}.
The proposed \ac{GMPPI} controller shows on average a 31\% reduction in position \ac{RMSE} over the existing \ac{MPPI} implementation~\cite{minarik2024} for agile trajectories.
In the ''Hover`` trajectory, a 97\% and 98\% reduction of maximum velocity and acceleration respectively has been achieved.
\ac{GMPPI} exhibits a 20\% higher position \ac{RMSE} than the standalone SE(3) controller.
This is an effect introduced by the \ac{GMPPI} cost function, which must consider all rollout trajectories, including the ones dedicated to exploring the space for obstacle-free areas.
In return for this small decrease in precision, the ability to avoid obstacles on the controller level is gained, which is critical for avoiding the risk of controller short-cutting, especially in agile flights.
The \ac{RMSE} of heading has been reduced compared to the previous \ac{MPPI} implementation~\cite{minarik2024} by an average of almost 88\%.
These improvements are driven by features presented in \refsec{sec:methodology}.
In particular, they highlight the synergy between SE(3) and \ac{MPPI} control described in \refsec{subsec:se3}.
The combination of these methods yields an obstacle-aware controller retaining most of the precision of the standalone SE(3) controller.

\begin{table}[t]
  \centering
  \caption{Results of reference tracking experiments.}
  \vspace{-1.2em}
  \label{tab:traj_results}
  \includegraphics[width=0.9\linewidth]{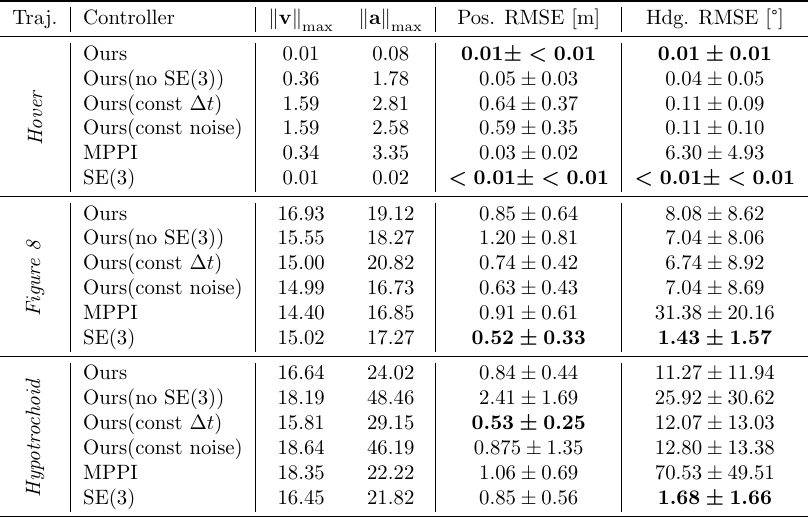}
  \vspace{-1.5em}
\end{table}


\subsection{Obstacle Avoidance}\label{subsec:res_obs}

\begin{figure}[t]
  \centering
  \includegraphics[width=1.0\linewidth]{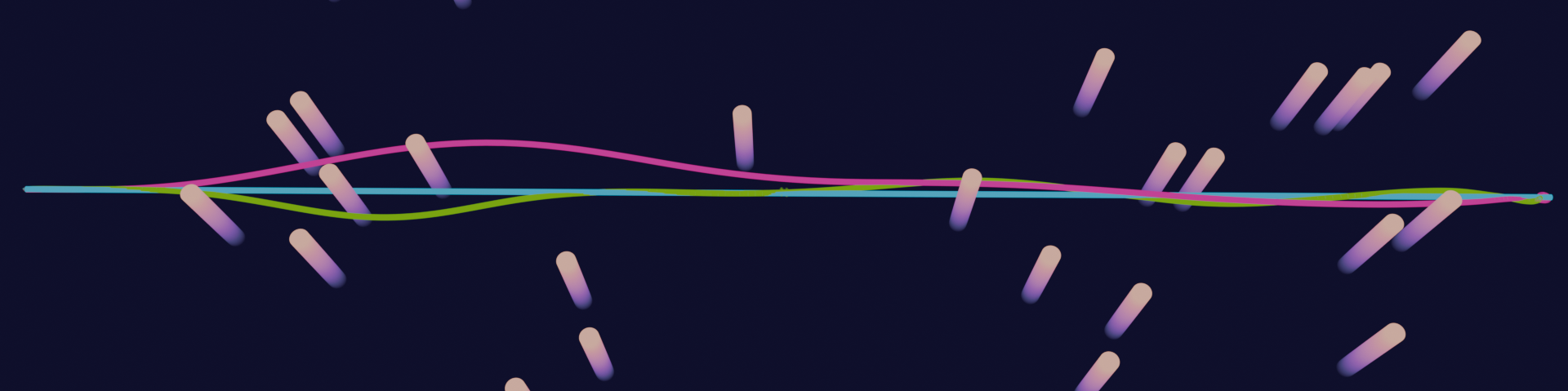}
  \vspace{-2.0em}
  \caption{Sample realization of a forest at a density $\delta = \nicefrac{1}{25}$, with the reference trajectory in blue and actual trajectories taken by the \ac{UAV} when controlled by \ac{GMPPI} at $5~\si{\m\s^{-1}}$ and $10~\si{\m\s^{-1}}$ in green and red respectively. \ac{GMPPI} balances the precision and smoothness while avoiding obstacles, which leads to slightly different trajectories at different flight speeds.}
  \label{fig:forest}
  \vspace{-1.0em}
\end{figure}

To verify obstacle avoidance capability, a series of flight experiments in ergodic forests \cite{karaman2012} was carried out. 
Each forest contained trees of diameter $0.6\si{\m}$ positioned according to a Poisson point process with a given density $\delta = \frac{1}{25}~\text{tree}\cdot\si{m^{-2}}$, as in \cite{loquercio2021}, \cite{ren2022} and \cite{zhang2024} to allow a fair comparison with the methods described therein.
A straight-line reference trajectory goes $40\si{\m}$ through the forest at a given speed ranging from $3\si{\m\s^{-1}}$ to $13\si{\m\s^{-1}}$.
A forest realization is shown in \reffig{fig:forest}.

\begin{figure}[b]
  \centering
  \vspace{-0.5cm}
  \includegraphics[width=0.85\linewidth]{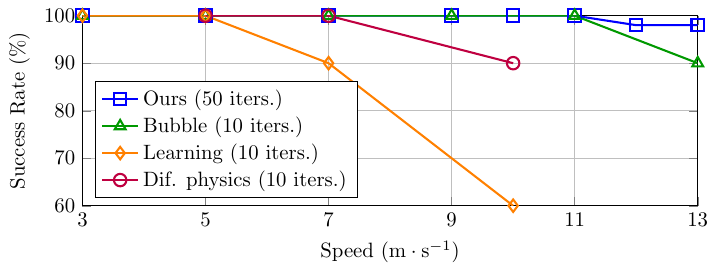}
  \vspace{-0.3cm}
  \caption{Graph of success rates for different controllers in the Poisson forest in simulated experiments. Bubble results are from~\cite{ren2022}, Learning results are from~\cite{loquercio2021} and Dif. physics results are from~\cite{zhang2024}.}
  \label{fig:graph_forest}
\end{figure}

The results of the simulated experiments are shown in \reffig{fig:graph_forest}, where the ``Bubble'' results are from \cite{ren2022}, the ``Learning'' results are from \cite{loquercio2021}, and the ``Dif. physics'' results are from~\cite{zhang2024}.
In the given forest density, the proposed \ac{GMPPI} controller achieves a 100\% success rate up to a speed of \SI{11}{\m\per\s}, surpassing all learning-based methods, with failures at higher speeds occurring at challenging sections of the forest with higher local densities.
At \SI{13}{\m\per\s}, the proposed method has 80\% less unsuccessful attempts than the Bubble planner~\cite{ren2022}.
It is worth noting that the Bubble planner uses a sensor with a range of \SI{8}{\m}, while the \ac{GMPPI} controller had the rollouts length limited to \SI{10}{\m}.

\subsection{Real-world experiments}

\begin{figure}
  \centering
  \includegraphics[width=1.0\linewidth]{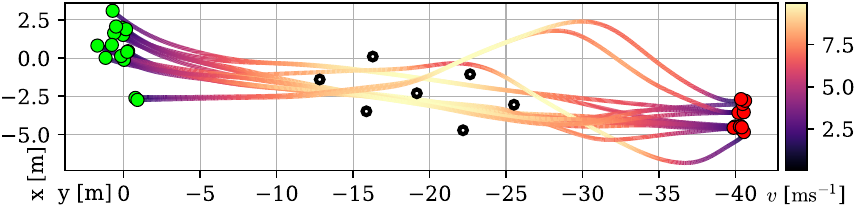}
  \vspace{-2.0em}
  \caption{Real-world flight using the \ac{GMPPI} controller. Filled lines represent UAV trajectories; black dots denote obstacles, and green/red dots mark the start/end points. Obstacle density is $\delta = \frac{1}{12} \, \mathrm{pylons} \, \cdot \, \si{\per\square\m}$ with a maximum reference velocity of $\SI{10}{\m\per\s}$.\label{fig:realworld_flight}}
\end{figure}

\begin{table}[t]
\centering
\scriptsize
\caption{Minimum Distance to Obstacles - Statistics}
\label{tab:pylon_stats}
\vspace{-1.2em}
\resizebox{\columnwidth}{!}{
\begin{tabular}{l|c|c|c|c}
\toprule
Velocity & Count & Mean [m] & Std. Deviation [m] & Worst-case [m] \\
\midrule
$\SI{10}{\m\per\s}$ & 15 & 0.38 & 0.14 & 0.18 \\
$\SI{7}{\m\per\s}$ & 5 & 0.34 & 0.07 & 0.24 \\
$\SI{3}{\m\per\s}$ & 6 & 0.56 & 0.09 & 0.44 \\
\bottomrule
\end{tabular}
}
\vspace{-2.0em}
\end{table}

Experiments with collision-free trajectories at speeds of up to \SI{17}{\m\per\s} and accelerations of up to \SI{35}{\meter\per\second\squared} were conducted to validate controller performance.
In a figure~8-shaped trajectory with maximum speed of \SI{14}{\m\per\s} and acceleration of \SI{26}{\meter\per\second\squared}, \ac{GMPPI} achieved a position \ac{RMSE} of \SI{0.69}{\m}, showing no performance loss compared to simulation.
To verify obstacle avoidance capability in the real world, a total of $42$ flights were conducted through an obstacle course composed of seven pylons, which simulated trees, but allowed preserving localisation capability based on RTK GPS.
A virtual ceiling was added to prevent the \ac{UAV} from flying over the obstacles.
Of the $42$ conducted flights, $26$ were carried out using a straight line reference trajectory through a pylon configuration shown in \reffig{fig:intro} with a density of $\delta = \frac{1}{12} \, \mathrm{pylons} \, \cdot \, \mathrm{m}^{-2}$.
These flight paths and the obstacle layout are shown in \reffig{fig:realworld_flight}.
The remaining experiments were carried out at speeds $\SI{3}{\m\per\s}$ and $\SI{7}{\m\per\s}$.
No collision has been recorded and the statistics of minimum distances to obstacles are shown in \reftab{tab:pylon_stats}.


\vspace{-0.5em}

\section{Conclusion}\label{sec:conclusion}

We presented \ac{GMPPI}, a controller for precise agile tracking with obstacle avoidance.
To our knowledge, it is the first to combine sampling-based \ac{MPPI} with geometric SE(3) control.
The ability to track agile trajectories was tested in simulation, with speeds up to $16~\si{\m~\s^{-1}}$ and acceleration in excess of $22~\si{\m~\s^{-2}}$. 
The results show that the controller achieves similar position \ac{RMSE} to the SE(3) controller on agile trajectories.
At the same time, our controller outperformed learning-based methods as well as the state-of-the-art obstacle-aware Bubble planner in obstacle avoidance capabilities.
In real-world experiments, \ac{GMPPI} showed no decrease in performance in following agile trajectories, and it was able to avoid obstacles at speeds of up to \SI{10}{\m\per\s}.

\vspace{-0.8em}

\bibliographystyle{IEEEtran}
\bibliography{main}

\vfill

\end{document}

%% file: assets/tikz/intro_figure.tex
\subfloat{\begin{tikzpicture}
    \node[anchor=south west,inner sep=0] (a) at (0,0) {
    \includegraphics[trim={0 3cm 0 3cm}, clip,width=1.0\linewidth]{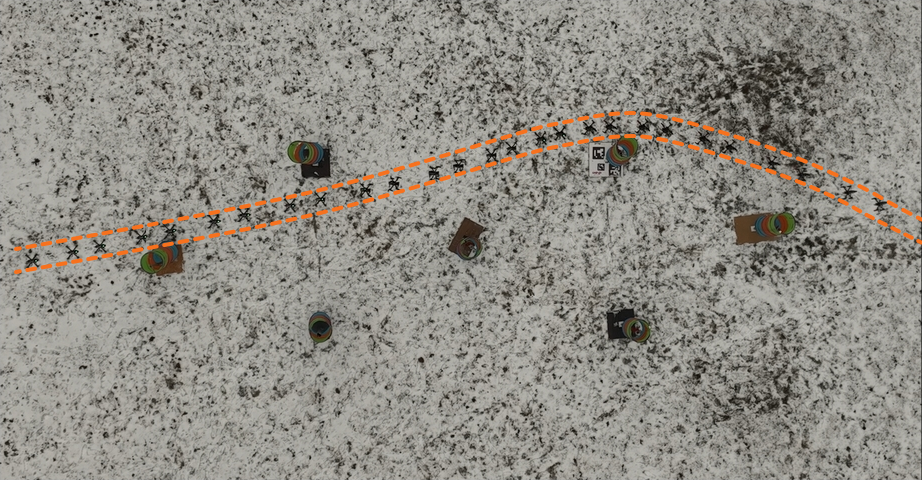}};
    \draw[white, very thick] (a.south west) rectangle (a.north east); 
    \begin{scope}[x={(a.south east)},y={(a.north west)}]
    \end{scope}
    \fill[white, opacity=0.8] (0.4, 0.4) circle (0.3cm);
    \draw (0.4,0.4) node [text=black, opacity=1] {\small (a)};
\end{tikzpicture}} \\
\vspace{-0.35cm}
\subfloat{\begin{tikzpicture}
  \node[anchor=south west,inner sep=0] (a) at (0,0) {\includegraphics[clip, trim={2.0cm 1.5cm 2.0cm 1.8cm},width=0.509\linewidth]{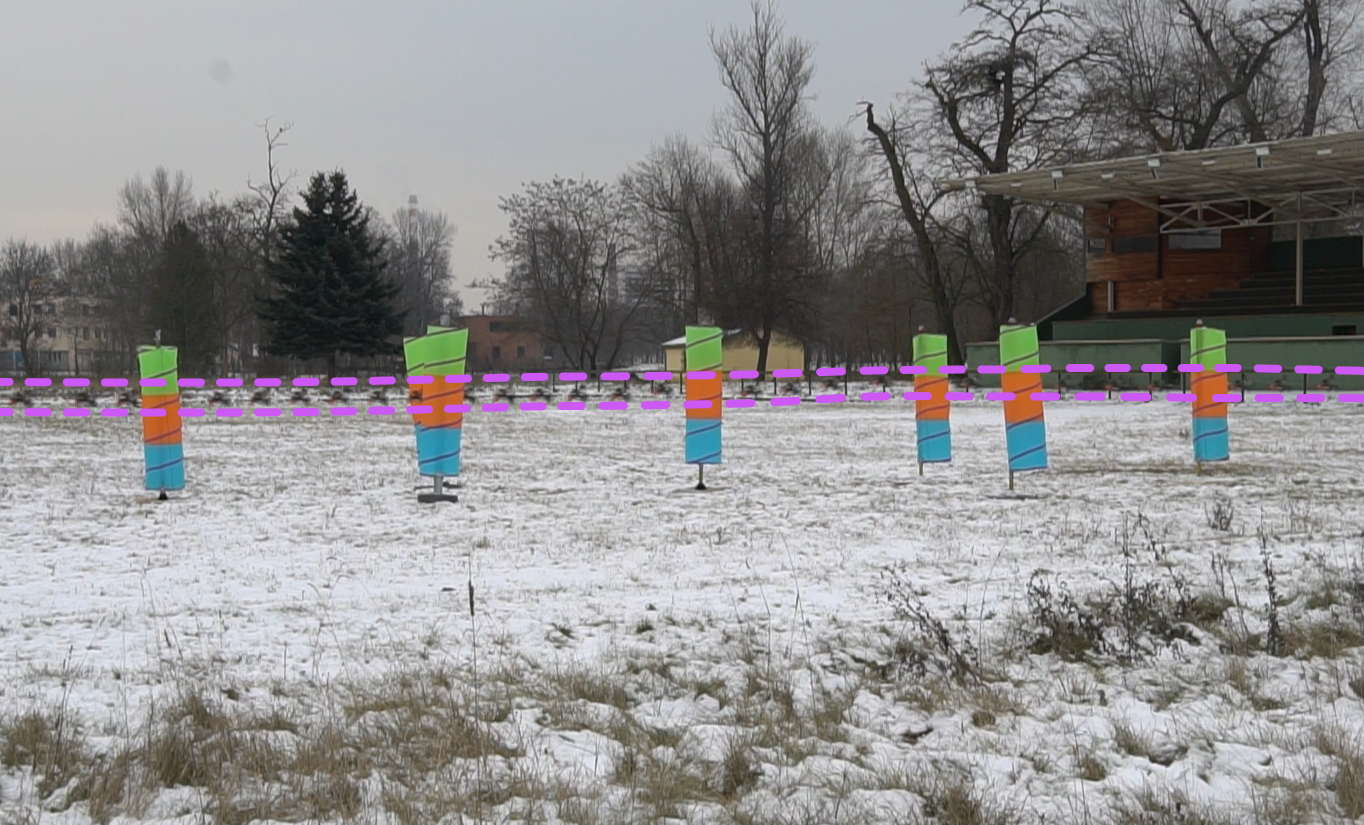}};
    \draw[white, very thick] (a.south west) rectangle (a.north east); 
    \begin{scope}[x={(a.south east)},y={(a.north west)}]
    \end{scope}
    \fill[white, opacity=0.8] (0.4, 0.4) circle (0.3cm);
    \draw (0.4,0.4) node [text=black, opacity=1] {\small (b)};
\end{tikzpicture}}
\subfloat{\begin{tikzpicture}
    \node[anchor=south west,inner sep=0] (a) at (0,0) {\includegraphics[clip, trim={0 3cm 0 0},width=0.485\linewidth]{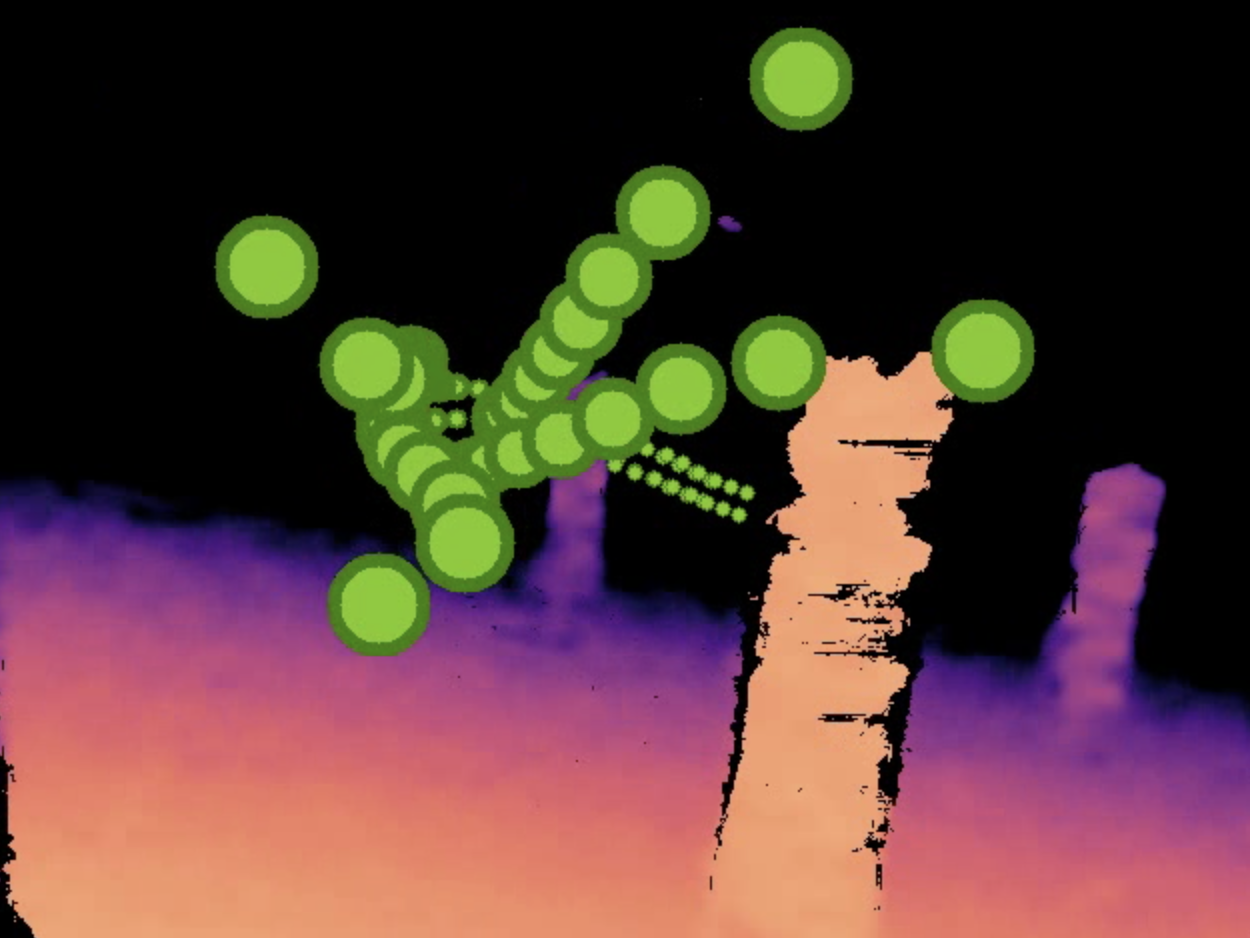}};
    \draw[white, very thick] (a.south west) rectangle (a.north east); 
    \begin{scope}[x={(a.south east)},y={(a.north west)}]
    \end{scope}
    \fill[white, opacity=0.8] (0.4, 0.4) circle (0.3cm);
    \draw (0.4,0.4) node [text=black, opacity=1] {\small (c)};
\end{tikzpicture}}